\documentclass{article}
\usepackage{iclr2018_conference,times}
\iclrfinalcopy

\usepackage[utf8]{inputenc} % allow utf-8 input
\usepackage[T1]{fontenc}    % use 8-bit T1 fonts
\usepackage{hyperref}       % hyperlinks
\usepackage{url}            % simple URL typesetting
\usepackage{booktabs}       % professional-quality tables
\usepackage{amsfonts}       % blackboard math symbols
\usepackage{nicefrac}       % compact symbols for 1/2, etc.
\usepackage{microtype}      % microtypography

\usepackage{ifthen} % A felteteles vezerleshez.
\usepackage{color}
\usepackage{makecell}
\usepackage{graphics}
\usepackage{amsmath}
\usepackage{amssymb}
\usepackage{mathtools}
\usepackage{multirow}
\usepackage{algorithm}
\usepackage{amsthm}
\usepackage{float}
\usepackage{comment}
\usepackage{booktabs}
\usepackage[title]{appendix}

\newboolean{showComments}
\setboolean{showComments}{false}
\ifthenelse
{\boolean{showComments}}
{
\newcommand{\wrk}[1]{\textcolor[rgb]{.0,.6,.0}{(\textsl{#1})}}
}
{
\newcommand{\wrk}[1]{}
}

\DeclarePairedDelimiter{\norm}{\lVert}{\rVert}
\DeclarePairedDelimiterX{\scal}[2]{\langle}{\rangle}{#1, #2}

\newcommand{\gradx}{\frac{\partial}{ \partial x}}
\newcommand{\softmax}{\mathop{\mathrm{softmax}}}

\newtheorem{theorem}{Theorem}
\newtheorem{claim}[theorem]{Claim}

\title{Gradient Regularization Improves Accuracy of Disciminative Models}

\author{D\'{a}niel Varga, Adri\'{a}n Csisz\'{a}rik, Zsolt Zombori \\
Alfr\'{e}d R\'{e}nyi Institute of Mathematics\\
Hungarian Academy of Sciences\\
Budapest, Hungary \\
\texttt{\{daniel,csadrian,zombori\}@renyi.hu} \\
}

\begin{document}

\maketitle

\begin{abstract}
Regularizing the gradient norm of the output of a neural network with respect to its inputs is a powerful technique, rediscovered several times. This paper presents evidence that gradient regularization can consistently improve classification accuracy on vision tasks, using modern deep neural networks, especially when the amount of training data is small. We introduce our regularizers as members of a broader class of Jacobian-based regularizers. We demonstrate empirically on real and synthetic data that the learning process leads to gradients controlled beyond the training points, and results in solutions that generalize well.

\end{abstract}

\section{Introduction}

Regularizing the gradient norm of a neural network's output with respect to its inputs is an old idea, going back to \emph{Double Backpropagation}~\citep{doublebp}. Variants of this core idea have been independently rediscovered several times since 1991~\citep{Sokolic, DataGrad, sobolev, wgan_gp}, most recently by the authors of this paper. Outside the domain of neural networks, \emph{Sobolev regularization}~\citep{NonparametricSparsityAndRegularization} is essentially the same concept, a special case of the very general, classic method of approximating functions in Sobolev space \citep{Gyorfi}. Smoothing splines~\citep{Wahba} are another important special case.

Most recent applications~\citep{Panasonic, Sokolic, DataGrad, RossDoshiVelez1711} focus on robustness against adversarial sampling~\citep{adversarial_samples}. Here we argue that gradient regularization can be used for the more fundamental task of increasing classification accuracy, especially when the training set is small. Calculating and implementing gradient regularization terms is made easy and fast by modern tensor libraries (see Appendix~A), making our proposed approach readily available for networks that are hundreds of layers deep.

Gradient regularization penalizes large changes in the output of some neural network layer, to enforce a smoothness prior. Our work explores a broad class of of Jacobian-based regularizers, providing a unified framework for various gradient regularization approaches. Part of our contribution is to demonstrate that this approach increases classification accuracy for models trained on relatively small datasets.
After introducing the general framework, we focus on the two most promising variants: 

\begin{itemize}
    \item \emph{Double Backpropagation}, which was discovered long ago, but its value as a regularizer on modern deep architectures has not yet been appreciated.
    \item \emph{Spectral Regularizer}, which is our own contribution and which seems to be the best variant for more complex datasets.
\end{itemize}

A possible objection against gradient regularization is that it focuses its attention on the metric of the input space, as opposed to the latent spaces emerging during training. However, recent success of techniques with the same limitation, such as \emph{mixup}~\citep{mixup} and \emph{Gradient Penalty}~\citep{wgan_gp}, weakens this objection. Nevertheless, overcoming this limitation is a promising future research direction, possibly by regularizing gradients with respect to hidden layers.

A more fundamental objection against minimizing the gradient norm at the training points is that this constraint can be satisfied locally, without any globally preferred behavior. We demonstrate through experiments on real and synthetic tasks that such locally optimal but globally unproductive solutions are hard to find with stochastic gradient descent. Instead, we obtain solutions that generalize well.

\section{Analysis}
\label{sec:analysis}

We consider feed-forward classifier networks with a loss function $L(x, y, \Theta) = M(f(x, \Theta), y) = M(\softmax (g(x,\Theta)), y)$, where $x$ is the input, $y$ is the one-hot encoded desired output, $f$ represents the network with a $\softmax$ layer on top, $\Theta$ are the network parameters and $M$ is the categorical cross-entropy function. The inputs and outputs of the $\softmax$ layer are called \emph{logits} and \emph{probabilities}, respectively. The central object of our investigation is the Jacobian of the logits $J_{g}(x) = \frac{\partial}{\partial x}g(x)$ and probabilities $J_{f}(x) = \frac{\partial}{\partial x}f(x)$ with respect to the inputs.

\subsection{Gradient Regularizaton Schemes}
\label{subsec:variants}

The general idea of penalizing large gradients can be applied in different variations, depending on where gradients are computed (logits, probabilities, loss term), with respect to what (inputs or some hidden activations), what loss function is used to create a scalar loss (in our investigation, we use the squared $L2$ norm). We summarize the most promising approaches: 

\paragraph{\emph{Double Backpropagation (DoubleBack)}~\citep{doublebp}} 
Also called \emph{DataGrad} after \citet{DataGrad}. Take the original loss term and penalize the squared $L2$ norm of its gradient. 
$$L_{DG}(x, y, \Theta) = L(x, y, \Theta) + \lambda \norm{(\frac{\partial}{\partial x}L(x, y, \Theta))}^2_2$$ 

Although not obvious from its definition, DoubleBack can be interpreted as applying a particular projection to the Jacobian of the logits and regularizing it. We prove this in Subsection~\ref{subsec:gradient_location}.

\paragraph{\emph{Jacobian Regularizer (JacReg)}~\citep{Sokolic}}
Penalize the squared Frobenius norm of the Jacobian of the softmax output (probabilities) with respect to the input.
$$ L_{JacReg}(x, y, \Theta) = L(x, y, \Theta) + \lambda \norm{J_{f}}^2_F $$

Symbolically computing the Jacobian is expensive as computation scales linearly with the number of output labels. To alleviate this, a layer-wise approximation is employed in \citet{Panasonic} and \citet{Sokolic}.

\paragraph{\emph{Frobenius Regularizer (FrobReg)}}
(Our contribution, also introduced by \citet{Giryes2018adversarial}.) Penalize the squared Frobenius norm of the Jacobian of the logits with respect to the input.
$$ L_{FrobReg}(x, y, \Theta) = L(x, y, \Theta) + \lambda \norm{J_{g}}^2_F $$

FrobReg only differs from JacReg in that the Jacobian is computed on the logits instead of the probabilities. Computation is equally expensive, however, diminishing gradients due to the softmax transformation are less of a concern.

\paragraph{\emph{Spectral Regularization (SpectReg)}}
(Our contribution.) Apply a random projection to the Jacobian of the logits, and penalize the squared $L2$ norm of the result:
$$ L_{SpectReg}(x, y, \Theta) = L(x, y, \Theta) + \lambda \norm{P_{rnd}(J_g)}^2_2 $$
    
where $P_{rnd}(J_g) = J_g^T r$ and $r \in \mathcal{N}(0, I_m)$.

When the projector is normalized onto the unit sphere (\emph{spherical SpectReg}), the norm of the random projection can be interpreted as a lower bound to the (hard to compute) spectral norm of the Jacobian. We show in Subsection~\ref{subsec:projected} that spherical SpectReg is an unbiased estimator of the squared Frobenius norm of the Jacobian. The same is true for the unnormalized variant, up to a constant scaling.~\footnote{We have not observed any empirical differences between the spherical and the unnormalized variants.}

%\paragraph{\emph{True label one-hot projection (OneHot)}}
%On the Jacobian of the logits, apply the projection:
%$$P_{True}(J_g) = J_{g}^T y$$ and penalize its $L2$ norm. This approach penalizes sudden changes in the true label's logit at the data points, but does not consider sudden changes in the other labels' logits. 

%We have found that OneHot's classification accuracy is similar to SpectReg's, but slightly below it, so in this paper we focus on SpectReg instead.

%\paragraph{\emph{Random one-hot projection (RandomOneHot)}}
%On the Jacobian of the logits, apply projection:
%$$P_{OneRnd} = J_g^T \sqrt{m} r$$ where $r$ is a uniformly chosen random basis vector, and penalize its $L2$ norm.~\footnote{The scaling factor $\sqrt{m}$ makes the covariance matrix the indentity.} 

%Similar to OneHot, this approach brings improvements compared to our baselines, but it lags behind SpectReg, so we do not deal with it in the rest of the paper.

\subsection{Starting from the linear case}
It is instructive to first consider the toy edge case when the neural network consists of a single dense linear layer. Here the weight matrix and the Jacobian coincide. Thus, for this network, FrobReg is identical to L2 weight decay and SpectReg is an estimator for weight decay.

The Frobenius norm is submultiplicative, and the gradient of the ReLU is upper bounded by 1. Thus, for a dense ReLU network the product of layer-wise weight norms is an upper bound for the FrobReg loss term. Applying the inequality of arithmetic and geometric means, we can see that the total weight norm can be used to upper bound the FrobReg loss term. Penalizing the Frobenius norm of the Jacobian seems to be closely connected to weight decay, however, the former is more targeted to the data distribution.

The Frobenius norm also serves as a proxy for the spectral norm, as they are equivalent matrix norms. For multi-valued functions, calculating, or even approximating the spectral norm is infeasible. Hence, Frobenius norm regularization is a reasonable approach to enforce a Lipschitz-like property.
Note, however, that the example of L1 vs. L2 weight regularization reminds us that optimizing different regularization terms can lead to very different behavior even when they are equivalent norms.

\subsection{Choosing where the gradient is computed}
\label{subsec:gradient_location}

Penalizing gradients computed at different vectors (logits, probabilities or loss) leads to different regularizing effects, worth comparing theoretically.

\paragraph{Logits vs. probabilities}
The Jacobian matrix of the softmax transformation is $J_{softmax} = (p \mathbf{1}^T) \odot (I - p \mathbf{1}^T)^T$, where $p$ is the probability vector and $\mathbf{1}$ denotes the all-one vector. Consequently:
\begin{align}
    J_f = J_{softmax} J_g = \left[ (p \mathbf{1}^T) \odot (I - p \mathbf{1}^T)^T \right] J_g  \label{eq:prob_jacobian}
\end{align}
One can easily verify that as $p$ tends to a one-hot vector, $J_{softmax}$ tends to the zero matrix. This results in a vanishing effect: once the network starts to converge, JacReg disappears, while FrobReg and SpectReg does not.

JacReg does not punish big changes in the logits when they do not affect class probabilities. This can lead to less robust predictions. Indeed, as our experiments confirm, FrobReg and SpectReg consistently outperform JacReg.

\paragraph{Probabilities vs. loss}
The Jacobian of the cross-entropy function is $J_{xent} = - \left[ \frac{y}{p}\right]^T$, hence:
\begin{align}
    J_{loss} & = J_{xent} J_f = \left[- \frac{y}{p} \right]^T J_f
\end{align}
Assuming that $y$ is a one-hot vector, the gradient at the loss is obtained from the Jacobian at the probabilities by selecting row $t$ of $J_f$ corresponding to the true label and multiplying it with $-\frac{1}{p_t}$. Most importantly, we no longer optimize the full Jacobian $J_f$, only a single row. 

\paragraph{Logits vs. loss} Composing the softmax and cross-entropy Jacobians, we obtain:
\begin{align}
    J_{loss} & = J_{xent} J_{softmax} J_g  = \left[ - \frac{y}{p}   \right]^T \left[ (p \mathbf{1}^T) \odot (I - p \mathbf{1}^T)^T \right] J_g \\
    J_{loss} & = \left[ - \frac{1}{p_t} \right] \left[ (p_t \mathbf{1}^T) \odot (y^T - p^T) \right] J_g  \\
    J_{loss} & = (p - y)^T J_g \label{eq:loss_jacobian}
\end{align}

Equation~\ref{eq:loss_jacobian} shows that calculating the gradient at the loss corresponds to applying a particular projection to the Jacobian of the logits, revealing a close connection between DoubleBack and SpectReg. DoubleBack applies a projection in a particular direction, while SpectReg selects a uniformly random direction, hence aims to control the gradients in all directions.

As both our analysis and experiments show, the gradient norms computed after the softmax layer quickly vanish during training, which indicates that the weights of such regularizers (JacReg, DoubleBack) require careful tuning. Indeed, our experiments confirm this hypothesis. Regularization based on the logits (FrobReg, SpectReg) is significantly more robust to choice of hyperparameter.

\subsection{Computationally efficient regularization of the Jacobian}
\label{subsec:projected}

Some of the variants presented in Subsection~\ref{subsec:variants} compute gradients on a vector (FrobReg, JacReg), while others first apply a projection (DoubleBack, SpectReg) and hence we compute the gradient of a single scalar. Due to the linearity of the gradient, both scenarios can be interpreted as regularizing the Jacobian matrix. If our regularizer is of the form $\norm{J_g^T w}^2$, then
\begin{align*}
\norm{\gradx \scal{g(x)}{w}} = \norm{(\gradx g(x))^T w}
\end{align*}

While in theory the projection should yield a negligible increase of computational burden, current tensor libraries all employ backwards automatic differentiation, which means that they do not reuse shared intermediate results when calculating the vector gradients, i.e. the rows of the Jacobian. Consequently, FrobReg/JacReg is vastly slower and vastly more memory-hungry than their peers that use projection, the performance ratio scaling with the number of class labels.
 
The following Claim is an easy consequence of the linearity of expectation:

\begin{claim}
    If $R$ is a distribution of row vectors with covariance matrix $I_m$, then $E_{r \sim R}[\norm{r J}^2_2] = \norm{J}^2_F$.
\end{claim}

%\begin{proof}
%    $$ E[\norm{r J}^2_2] = E[\Tr(J^T r^T r J)] = \Tr( J^T E[r^T r] J) = \Tr(J^T J) = \norm{J}^2_F $$
%\end{proof}

Consequently, spherical SpectReg is an unbiased estimator of the squared Frobenius norm of the Jacobian and so is SpectReg, up to constant scaling. 

%Besides the computational speedup, projections can introduce beneficial extra regularization.

%\subsection{Continuity violated}
%
%It is worth noting that for the important class of piecewise linear activation functions, gradient-based regularizers violate an important standard assumption: that the loss is a continuous function of the network parameters. $J_g(x)$ is a piecewise constant function of the inputs. It is also a piecewise constant function of the biases, implying that the biases of the network do not get a gradient update from the gradient-based loss terms. More importantly, loss is a non-continuous, piecewise multilinear function of the weights. Nevertheless, we have not observed optimization issues that could be attributed to non-continuity.

%Let us now assume using ReLU activations for simplicity. For a fixed input $x$ and a fixed $\Theta$, let us call the set of neurons with positive output the network's firing pattern for the given $x$ and $\Theta$. Gradient regularization terms are continuous on subsets of the parameter space where the firing pattern is constant. We speculate that the reason non-continuous gradient regularization loss terms can successfully be optimized using gradient descent is that they are ``close to continuous'', in the sense that a single gradient update does not cause a drastic change in the firing pattern for a given input.

\section{Related Work}
\label{sec:related}

%Our inspiration for using symbolically computed gradient norms to regularize a classifier network came from the Gradient Penalty technique by \citet{wgan_gp}, who employ symbolically computed gradient norms to train Wasserstein GANs. However, after obtaining the first, promising implementation of our method, our literature search gradually unearthed a large amount of earlier incarnations of the core idea. This literature search was made harder by the fact that the citation graph between these results is quite sparse, with many independent rediscoveries.

The idea to control gradient norms with respect to the inputs first appeared in \citet{doublebp} who called it \emph{Double Backpropagation}. However, due to slow hardware and a missing technical apparatus for symbolic differentiation at that time, they only evaluated the idea on networks with two hidden layers. 
The \emph{TangentProp} variant was introduced in \citet{tangentprop} where the magnitude of change is controlled only in some input space directions, corresponding to selected invariances. This, for example, can be used to promote rotational invariance by using the direction of infinitesimal rotation in pixel space.

%Later rediscoveries of the core idea demonstrate variation in implementation details and choice of applications. % Below we review the instances that we are aware of and compare our results to each of these.

\paragraph{Jacobian Regularization}

The \emph{Jacobian Regularizer} introduced by \citet{Sokolic} works by regularizing the Frobenius norm of the Jacobian of the softmax output with respect to the input. For smaller networks the authors symbolically compute the Jacobian. For larger networks they employ a Frobenius norm based regularization for each layer individually, for performance reasons. Such layer-wise regularization of the Frobenius norm was also employed in \citet{Panasonic}. 

%The computation of the symbolic Jacobian Regularizer roughly requires $m$ parallel computations of the forward pass, where $m$ is the number of output labels. The alternatives we present are significantly less expensive computationally while their accuracy is higher on our tasks.

%One difference between the Jacobian Regularizer and our \emph{Spectral Regularizer} is that we work with the Jacobian of the logits while \citet{Sokolic} work with the Jacobian of the softmax output. We experimentally compare these approaches and conclude that for our tasks the logits are the more appropriate choice. This can be justified theoretically by noting that diminishing gradients (with respect to weights) are more of a concern when a sigmoid (or in our case, softmax) is applied to the output.

\paragraph{Robustness to adversarial examples}

The Double Backpropagation formula was rediscovered by \citet{DataGrad} who named it \emph{DataGrad}. However, the authors do not actually implement the formula, rather, a finite difference approximation is used: after finding adversarial samples by gradient descent, they penalize large changes in the loss function between the data point and its adversarially perturbed version. This approximation improves robustness to adversarial sampling. Recently, \citet{RossDoshiVelez1711} use symbolically computed Double Backpropagation to improve model \emph{interpretability} as well as robustness against adversarial noise. 

%Neither of these works report classification accuracy increases.

\paragraph{Sobolev training}
A generalized form of the Double Backpropagation idea is when an oracle gives information about the gradients (or even the Hessian) of the target function, and we incorporate this into a loss term. This is investigated in \citet{sobolev}, focusing on \emph{distillation} and \emph{synthetic gradients}, applications where such an oracle is available. 
In contrast to this approach, we demonstrate the counter-intuitive fact that in the absence of such an oracle, simply pushing gradient norms toward zero at the data points can already have a beneficial regularization effect.

\paragraph{Spectral norm regularization}
Concurrent with our work \citet{spectral_norm_regularization} enforce smoothness by penalizing an upper bound to the spectral norm of the neural network at all points of the input space. The price of this global effect is that the upper bound is potentially very far from the actual spectral norm. Our approach, on the other hand, focuses on the data manifold and provides input specific regularization effect.

\paragraph{Sensitivity and generalization}
Very recently, \citet{novak2018sensitivity} perform a large scale empirical study that shows correlation between sensitivity to input perturbations and generalization. They find that for networks that generalize well, the Frobenius norm of the Jacobian tends to have smaller norm. Their findings strengthen the motivation for gradient regularizaton.

\section{Experiments}
\label{sec:experiments}

Our experiments compare gradient-based regularization with various other methods in order to position this technique on the landscape. Our results show that SpectReg and DoubleBack consistently increase accuracy of baselines on a wide variety of image classification datasets when the amount of training data is restricted. DoubleBack performs better on MNIST, however, on more complex datasets SpectReg takes the upper hand.

We present results on MNIST, CIFAR-10, CIFAR-100, TinyImageNet-200 and a simple synthetic dataset. Implementation details can be found in Appendix~B. Optimal hyperparameters of the particular experiments can be found in Appendix~C.

We emphasize that our experiments use reduced training sets as we found that gradient regularization is particularly useful in such scenarios. By \emph{small MNIST} and \emph{small CIFAR-10 / CIFAR-100} we refer to these datasets restricted to 200 randomly selected training points per class. We also run experiments on TinyImageNet-200~\footnote{https://tiny-imagenet.herokuapp.com}, a standard reduced version of the ImageNet dataset containing 500 training images per class, for a total of 100000 data points. None of our experiments employ data augmentation, to enforce the model to generalize from limited data.

An individual experiment consists of $10$ runs of the same setup, each time with a different randomly chosen training set. All reported numbers are the mean of these runs evaluated on 10000 test points, accompanied with the standard deviation in parentheses. Hyperparameters are tuned on 10000 validation points. The hyperparameters that are not explicitly discussed are all tuned for the baseline model, but our experiments suggest that they are good choices for the regularized models as well.

% In line with Section~\ref{sec:analysis}, $\lambda$ denotes the weight of the loss term associated with a particular regularizer.

\subsection{Weight decay and gradient regularization}

Weight decay is probably still the most widespread regularization method. On small MNIST, gradient regularization yields more accuracy gain than weight decay. For small CIFAR-10/100, weight decay is more important than gradient regularization, but the latter still brings significant benefit. In our exploratory experiments, DoubleBack slightly outperformed other variants on MNIST, however, we could not tune it on CIFAR to rival SpectReg. Overall, we found SpectReg to be the most robust regularizer and here we only present accuracy results for this variant.  Table~\ref{tab:mnist_wd_vs_nowd} shows our results. 

% DoubleBack {\bf 97.95 \small{(0.15)}}, 97.92 \small{(0.16)}

\begin{table}[htb]
    \caption{
        Gradient regularization alongside weight decay.
    }
    \centering
    \begin{tabular}{ l | l l | l l l }
        
        & \multicolumn{2}{l|}{Without weight decay} & \multicolumn{3}{l}{With weight decay} \\
        Dataset & Baseline & SpectReg & Baseline & SpectReg & WD \\
        \hline
        small MNIST & 97.25 \small{(0.22)} & 97.69 \small{(0.11)} & 97.40 \small{(0.15)} & {\bf 97.73} \small{(0.13)} & 0.0005 \\
        
        small CIFAR-10 & 48.27 \small{(0.82)} & 50.41 \small{(0.65)}  & 55.63 \small{(2.06)} & {\bf 59.24} \small{(1.48)} & 0.003 \\
        
        small CIFAR-100 & 37.81 \small{(0.33)} & 41.80 \small{(0.70)} & 49.56 \small{(2.96)} & {\bf 52.49} \small{(0.65)} & 0.003 \\
    \end{tabular}
    \label{tab:mnist_wd_vs_nowd}
\end{table}

\subsection{Gradient Regularization Compared with Dropout and Batch Normalization}

On small MNIST, we find that both DoubleBack and SpectReg outperform both Dropout~\citep{JMLR:v15:srivastava14a} and Batch Normalization~\citep{DBLP:conf/icml/IoffeS15}, two well established techniques. We observe further accuracy gain when Dropout and DoubleBack/SpectReg is combined. Table \ref{tab:mnist_1} summarizes our results. While in all other experiments the MNIST baseline includes dropout, here the baseline is only regularized with weight decay.

\begin{table}[htb]
    \caption{
        Comparison of Dropout and Batch Normalization versus DoubleBack and SpectReg on small MNIST. Both DoubleBack and SpectReg achieve higher accuracy than either Dropout or Batchnorm in itself. We obtain the best result by combining Dropout and DoubleBack.
    }
    \centering
    \begin{tabular}{ l l l l l }
%        \multicolumn{4}{c}{Dataset: MNIST, training set size 2000} \\
        & Dataset & NoGR & SpectReg & DoubleBack  \\
        \hline
        Baseline & small MNIST & 96.99 \small{(0.15)} & 97.59 \small{(0.13)} & 97.56 \small{(0.24)} \\
        Batchnorm & small MNIST & 96.89 \small{(0.23)} & 96.94 \small{(0.27)} & 96.89 \small{(0.22)} \\
        Dropout & small MNIST & 97.29 \small{(0.19)} & 97.65 \small{(0.14)} & {\bf 97.98} \small{(0.12)} \\
        
    \end{tabular}

    \label{tab:mnist_1}
\end{table}

% Since Batch Normalization is a crucial ingredient of large modern networks and its weakness is particular to the MNIST task, it is important to see how Gradient Regularization fares on large networks using Batch Normalization. We demonstrate in Subsection \ref{subsec:exp_ent} the usefulness of our methods on CIFAR-10, using a large residual network.

\subsection{Gradient Regularization versus Confidence Penalty}
\label{subsec:exp_ent}

One recently introduced regularization technique is \emph{Confidence Penalty (CP)}~\citep{pereyra}. It directly penalizes the negentropy of the softmax output, thus preventing model overconfidence at the training points. It is closely related to the technique of \emph{label smoothing}~\citep{LabelSmoothing}, and brings small but consistent accuracy improvements on many tasks. % Computationally, it is significantly cheaper than our methods.

We find that gradient regularization performs better than CP on small MNIST. They seem to perform orthogonal tasks and can be combined to obtain further improvements. Table~\ref{tab:mnist_2} shows our results.

\begin{table}[htb]
    \caption{
        Confidence Penalty versus SpectReg and DoubleBack on small MNIST. While all regularizers improve upon the baseline model, DoubleBack performs best. Furthermore, the benefits of gradient regularization and confidence penalty add up when used together.
    }
    \centering
    \begin{tabular}{l l l l l}
%        \multicolumn{4}{c}{Dataset: MNIST, train set size 2000} \\
        & Dataset & Baseline & SpectReg & DoubleBack  \\
        \hline
        No CP & small MNIST & 97.39 \small{(0.11)} & 97.79 \small{(0.12)} & 97.89 \small{(0.12)} \\
        CP & small MNIST & 97.57 \small{(0.15)} & 97.89 \small{(0.12)} & {\bf 98.07} \small{(0.09)} \\
        
    \end{tabular}

    \label{tab:mnist_2}
\end{table}

We also make these comparisons on small CIFAR-10 and small CIFAR-100, using a medium sized residual network (see Appendix~B) that is representative of modern vision models. As Table~\ref{tab:cifar_ent_dg_spect} shows, SpectReg clearly dominates. Note again that hyperparameters were tuned carefully.

\begin{table}[htb]
    \caption{
        Comparing SpectReg, DoubleBack and Confidence Penalty on small CIFAR-10/100.
    }
    \centering
    \begin{tabular}{l l l l l}
        Dataset & Baseline & SpectReg & DoubleBack & CP  \\
        \hline
        small CIFAR-10 & 55.63 \small{(2.06)} & {\bf 59.24} \small{(1.48)} & 57.45 \small{(1.79)} & 58.30 \small{(1.79)} \\
        small CIFAR-100 & 49.56 \small{(2.96)} & {\bf 52.49} \small{(0.65)} & 48.72 \small{(2.84)} & 51.04 \small{(1.29)} \\
        
    \end{tabular}
    \label{tab:cifar_ent_dg_spect}
\end{table}

\subsection{The effect of training set size}

\begin{figure}[htb]
    \centering
    \includegraphics[width=0.8\textwidth]{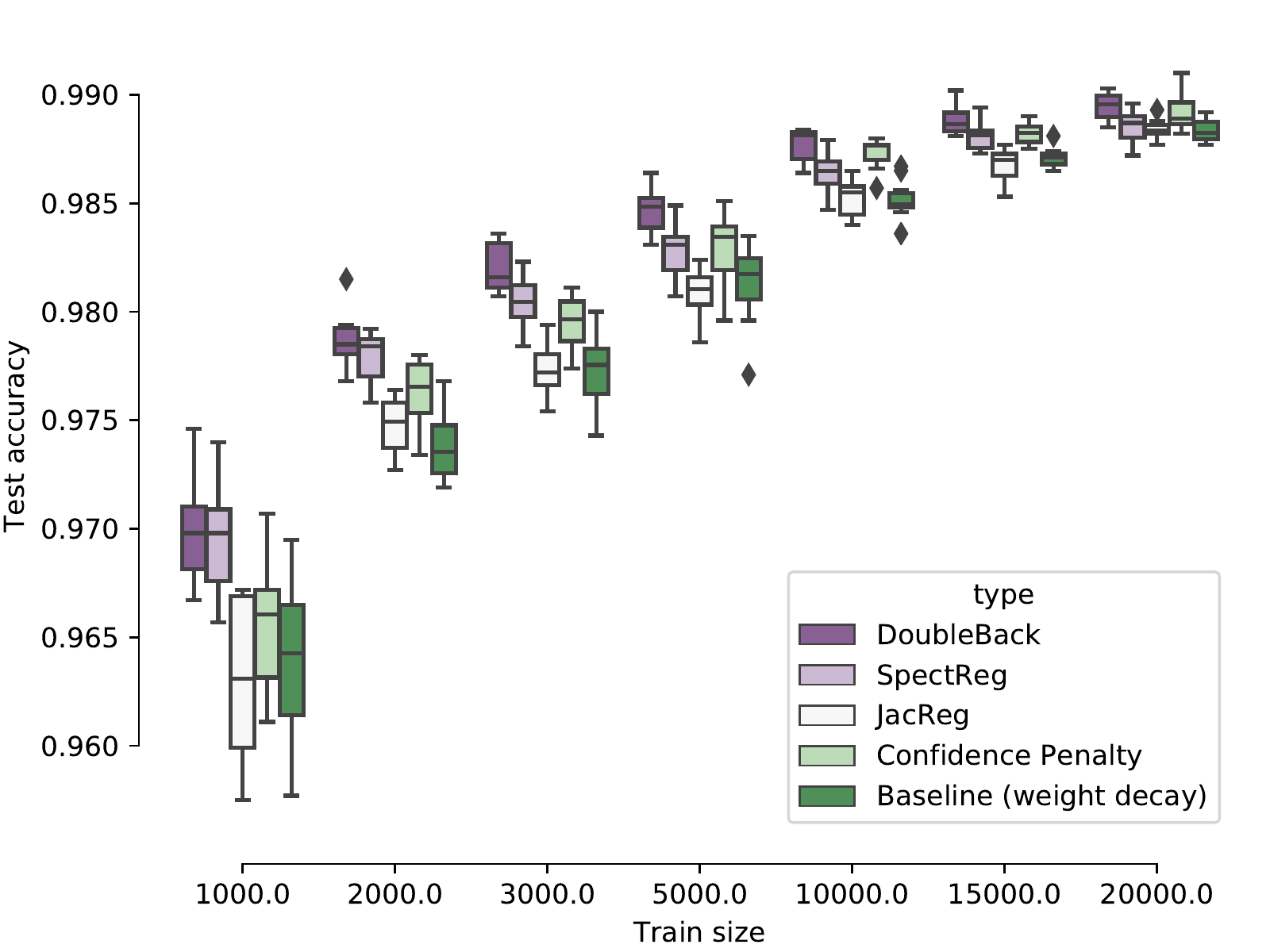}
    \centering
    \caption{
        Comparison of various regularization methods on MNIST using different training set sizes. DoubleBack (DataGrad) performs best consistently for all sizes.
    }
    \label{fig:mnist_4}
\end{figure}

The effect of regularizers is more significant for smaller training sets, however, as we show in Figure~\ref{fig:mnist_4} they maintain a significant benefit even for as much as $20000$ training points on MNIST. DoubleBack performs best for all sizes. SpectReg is better than CP for smaller sizes, but its advantage evaporates with more training data. JacReg is consistently worse than its peers.

\subsection{Scaling with the number of labels}
As Table~\ref{tab:tinyimagenet} shows, gradient regularization works well even for the significantly more complex TinyImageNet-200 dataset.

Note that in our various experiments the number of class labels range from 10 to 200, and the effect of gradient regularization does not fade with increased label count. This is noteworthy especially in the case of SpectReg, which is beneficial regardless of whether the number of matrix rows projected is 10 or 200.

\begin{table}[h!tb]
    \caption{
        Results on the TinyImageNet-200 dataset.
    }
    \centering
    \begin{tabular}{ l l l l l l }
         & & \multicolumn{2}{l}{Top-1 accuracy} & \multicolumn{2}{l}{Top-5 accuracy} \\
        Dataset & Baseline & SpectReg & Baseline & SpectReg \\
        \cmidrule(lr){1-2}\cmidrule(lr){3-4}\cmidrule(lr){5-6}
        TinyImageNet-200 & 44.62 \small{(0.58)}& {\bf 50.76} \small{(0.59)} & 70.20 \small{(0.58)}& {\bf 75.93} \small{(0.40)}  \\
        %TinyImageNet-200 & Yes & 55.32\small{(0.33)} & 55.58 \small{(0.90)} & 79.46 \small{(0.25)} & 79.38 \small{(0.65)}  \\
        %TinyImageNet-200 & Yes & 55.32\small{(0.33)} & 55.78 \small{(0.49)} & 79.46 \small{(0.25)} & 79.55 \small{(0.30)}  \\
    \end{tabular}
    \label{tab:tinyimagenet}
\end{table}

\subsection{Approximating the Frobenius norm of the Jacobian does not decrease accuracy}

As discussed in Subsection~\ref{subsec:projected}, directly minimizing the Frobenius norm of the Jacobian is an expensive operation. We demonstrate that applying a random projection to the Jacobian does not reduce the efficiency of the regularization while significantly reducing the computational burden.

We compare SpectReg, which applies a random projection to the Jacobian of the logits, JacReg which minimizes the full Frobenius norm of the Jacobian of the probabilities, and FrobReg which minimizes the full Frobenius norm of the Jacobian of the logits. We can see from Table \ref{tab:mnist_6} (and especially from Figure~\ref{fig:mnist_4}) that it is more beneficial to control the Jacobian on the logits, rather than on the probabilities. We also conclude that minimizing a random projection of the Jacobian does not lead to a loss in accuracy, compared with the full calculation of the Jacobian.

\begin{table}[htb]
    \caption{
        Comparing SpectReg, FrobReg and JacReg on small MNIST. SpectReg performs best, but the differences are small:
        % and we evaluated the statistical significance of the differences, obtaining the following p-values: SpectReg = FrobReg: $0.5984$, SpectReg = JacReg: $0.0169$, FrobReg = JacReg: $0.053$.
        after Bonferroni correction, the only statistically significant difference is between SpectReg and JacReg.
    }
    \centering
    \begin{tabular}{l l l l}
%         \multicolumn{3}{c}{Dataset: MNIST, training set size 2000} \\
         Dataset & SpectReg & JacReg & FrobReg \\
         \cmidrule(lr){1-1}\cmidrule(lr){2-4}
         small MNIST & {\bf 97.79\%} (0.12) & 97.63\% (0.15) & 97.76\% (0.13) \\
    \end{tabular}
    \label{tab:mnist_6}
\end{table}

\subsection{Local gradient control does not lead to pathological gradient landscape}

A reasonable objection to gradient regularization methods is that they control the gradients only in the training points. A highly overparametrized network is capable of representing a "step function" that is extremely flat around the training points and contains unwanted sudden jumps elsewhere. Such a solution results in low gradient loss with high actual gradient norms in some points. However, all our experiments indicate that the learning process of neural networks struggles to find such pathological solutions and instead smoothens the function on its whole domain. We emphasize that in all of our experiments except for TinyImagenet, the datasets are so small that the models are inevitably overfitting, yielding close to $100\%$ train accuracy. In such setting, it is remarkable that the gradient control does not overfit.

\begin{figure}[htb]
    \centering
    \includegraphics[width=0.8\textwidth]{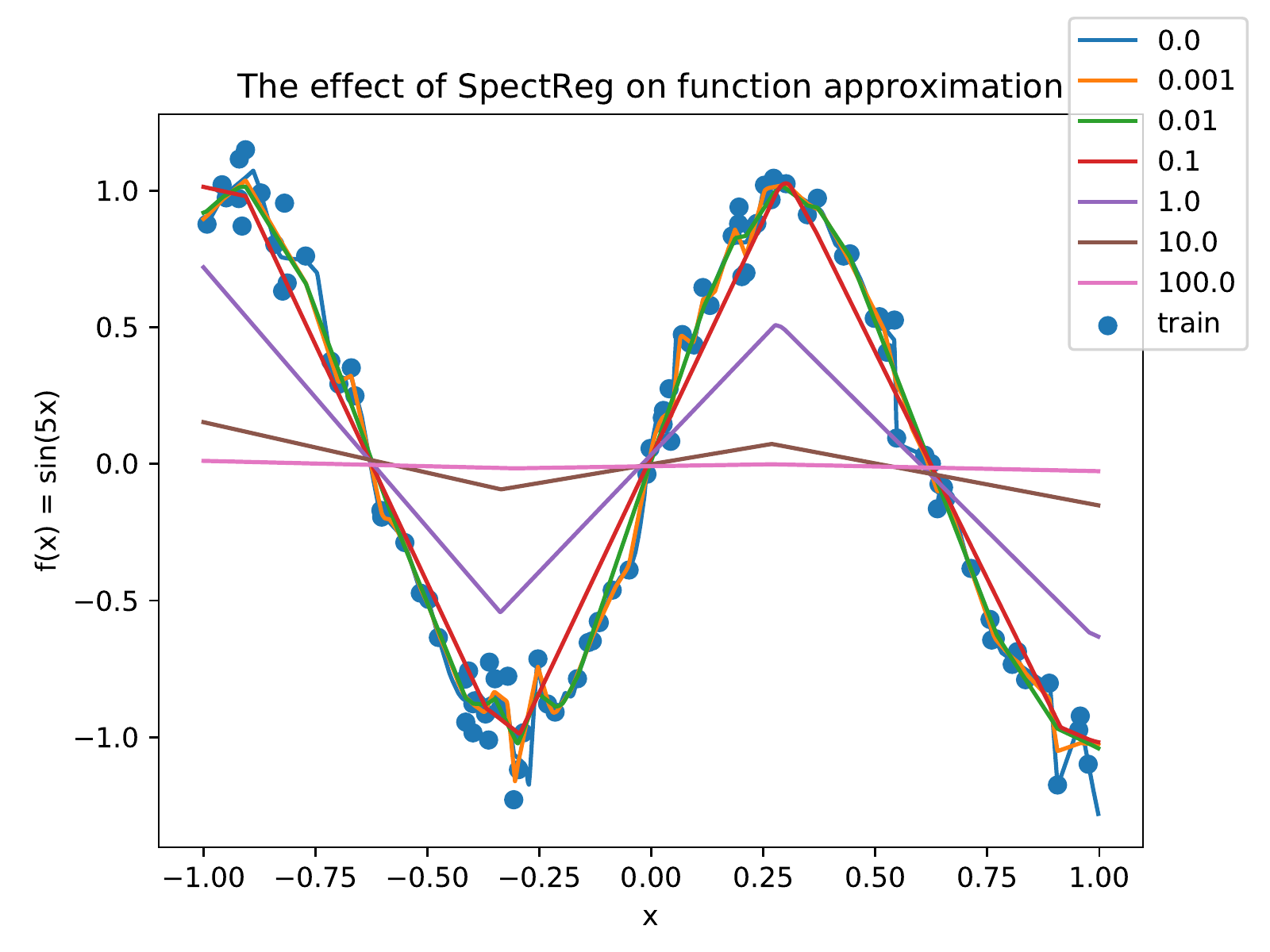}
    \centering
    \caption{
        Increasing the weight of the SpectReg regularizer on SIN forces the network to learn an increasingly flat function. Although the gradient is controlled only on $100$ training points, the whole manifold becomes smoother.
    }
    \label{fig:syn_spectreg}
\end{figure}

We demonstrate this on the SIN dataset. Note that our training set is small and noisy, so a good fit is likely to be curvy and steep. Figure~\ref{fig:syn_spectreg} shows our training set and the functions learned by the network regularized with SpectReg using different $\lambda$ weights.~\footnote{Using DoubleBack produces very similar graphs.} We observe that increasing $\lambda$ makes the output smoother, not only around the training points, but globally as well.

Note that the bandwidth of the network does not limit its ability to learn the aforementioned step-function. We verified this by training a modified version of the dataset: we have two datapoints $(x, y)$ and $(x+0.001, y)$ for each $(x, y)$ element of the original SIN. The network managed to reconstruct this many-plateau function, demonstrating that it is the optimization process rather than the network's bandwidth that prevents the model from reaching a step-function like solution. 

Somewhat counterintuitively, on the synthetic dataset, too large $\lambda$ weights manifest in learning sawtooth-shaped functions rather than step functions. In other words, SGD gets stuck at local optima with a very high SpectReg loss, exactly when the SpectReg loss term gets more emphasis.

\begin{figure}
    \centering
    \includegraphics[width=0.8\textwidth]{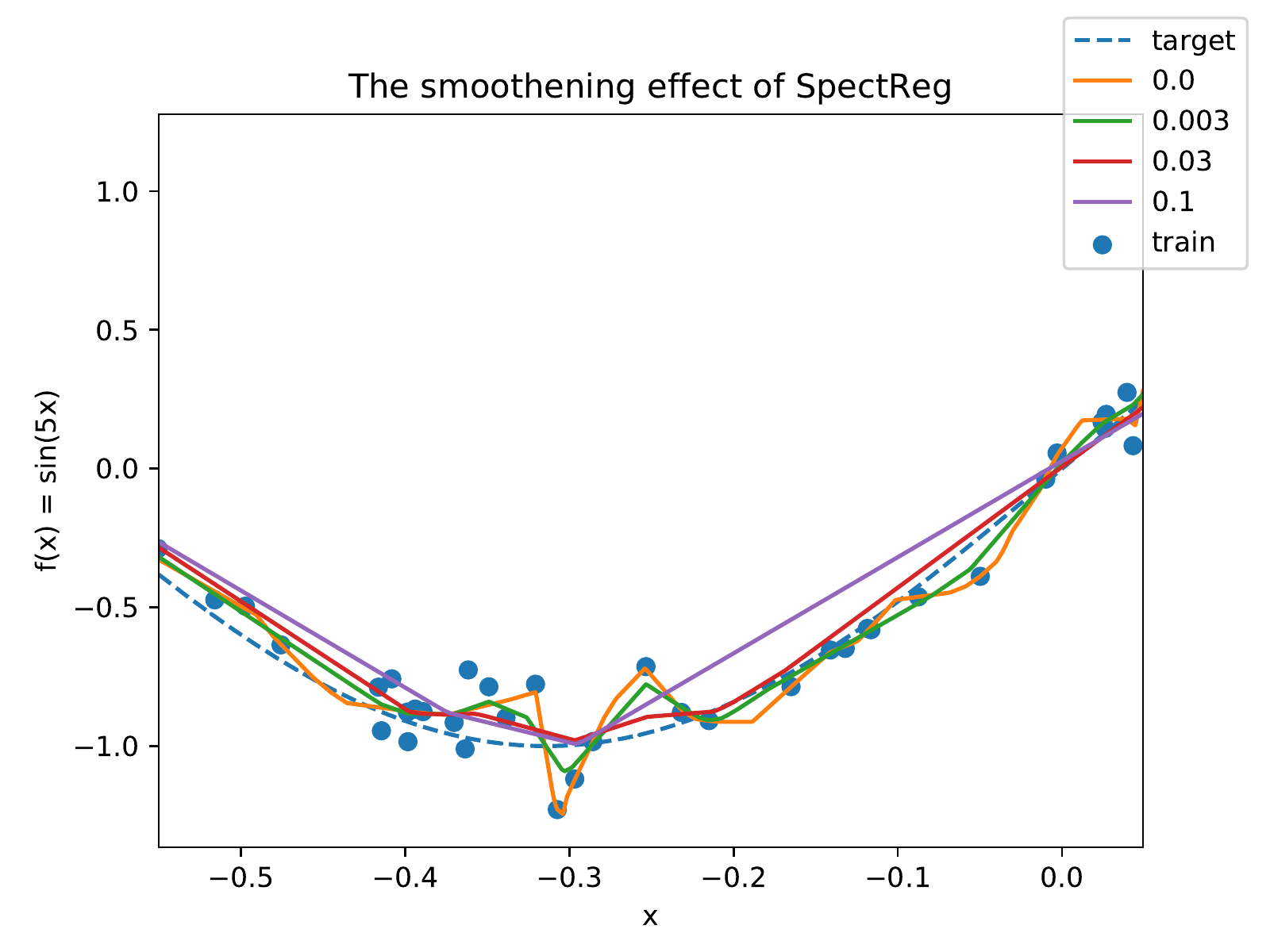}
    \caption{Zooming into a small portion of the domain, we observe that SpectReg allows for achieving a better fit to the target function.}
    \label{fig:synthetic_spectreg_zoomed}
\end{figure}

While too strong regularization degrades the approximation, a well tuned SpectReg yields much better fit to our target function. This is demonstrated in Figure~\ref{fig:synthetic_spectreg_zoomed}, where we zoom into a small portion of the domain. Table~\ref{tab:synthetic_spectreg} shows the mean squared error on the test set for various values of SpectReg $\lambda$. SpectReg can reduce the mean squared error of the baseline model by a factor of $10$.

\begin{table}[htb]
    \caption{
    Mean squared error (MSE) on SIN for various SpectReg weights. The optimal weight is $0.03$ which yields a reduction in MSE by a factor of $10$.
    }
    \centering
    \begin{tabular}{ l l l l l l l l l l l}
        SpectReg $\lambda$ & 0 & 0.001 & 0.003 & 0.01 & 0.03 & 0.1 & 0.3 & 1 & 3 & 10\\
        \hline
        MSE (1e-5) & 16.6 & 24.5 & 2.5 & 2.3 & \textbf{1.6} & 3.4 & 74.9 & 497.7 & 1264.4 & 2002.2 \\
        
    \end{tabular}
    \label{tab:synthetic_spectreg}
\end{table}

\section{Conclusion, Future Work}

Our paper presents evidence that gradient regularization can increase classification accuracy, especially for smaller training set sizes. We introduce \emph{Spectral Regularization} and after comparing it with other gradient regularization schemes, we find that SpectReg is the most promising variant.

Despite the fact that gradient control is applied only at the training points, we find that stochastic gradient descent converges to a solution where gradients are globally controlled. Even for very small training set sizes, the regularized models become smoother on the whole data manifold.

Below we list some of what we see as promising further research on gradient regularization:

\begin{itemize}
%    \item DoubleBack is sensitive to the choice of its coefficient. Understanding the relationship between the magnitude of the main loss and the DoubleBack loss could help scaling the magnitude, leading to more robust behavior.

    \item The experiments we present control gradients at labeled training points, but we note that SpectReg makes no use of labels and can be applied to unlabeled inputs in a semi-supervised setting.

    \item We considered gradients calculated with respect to inputs. In the future, we plan to investigate gradients calculated with respect to hidden activations.

    \item Our understanding of the counterintuitive interaction between gradient regularization and stochastic gradient descent is still limited. A full-blown theory may yet be out of sight, but it is reasonable to expect that well-chosen experiments on synthetic benchmarks will lead to important new insights.
\end{itemize}

\section*{Acknowledgement}
The research leading to these results has received funding from the European Research Council under the European Union’s Seventh Framework Programme (FP7/2007-2013) / ERC grant agreement 617747. The research was also supported by the MTA Rényi Institute Lendület Limits of Structures Research Group. We thank Csaba Szepesvári and Christian Szegedy for helpful discussions.

\medskip

\bibliography{main}
\bibliographystyle{iclr2018_workshop}

\newpage

\begin{appendices}

\section{Tensorflow implementation of DoubleBack and SpectReg}

Calculating gradient regularization terms is made easy and fast by modern tensor libraries.  Figure~\ref{fig:code_snippet} shows the full Tensorflow implementation of the two regularizers that we focus on.

\begin{figure}[htb]
\begin{verbatim}
doubleback_loss = tf.reduce_sum(
    tf.square(tf.gradients(loss, [x])[0]), axis=1)

gradients = tf.gradients(
    logits * tf.random_normal((OUTPUT_DIM,)), [x])[0]
spectreg_loss = tf.reduce_sum(tf.square(gradients), axis=1)
\end{verbatim}
\caption{TensorFlow implementations for the DoubleBack and SpectReg regularizers}
\label{fig:code_snippet}
\end{figure}

\section{Implemented networks}

For each regularizer, dataset and training set size, the optimal weight is selected based on a grid search from the following intervals: DataGrad $[0.0003, 100]$, SpectReg $[0.0001, 1]$, JacReg $[0.0001, 1]$, CP $[0.0001, 1]$. All other hyperparameters are tuned by gridsearch for the baseline model, but our experiments suggest that they are good choices for the regularized models as well.

%We find that these regularizers are not sensitive to their weight parameter, except for DataGrad, which requires more careful tuning. This is probably related to the fact that DataGrad vanishes as the network achieves a good fit on the training set. One needs to strike a careful balance to ensure that DataGrad vanishes neither too quickly nor too slowly.

\subsection{MNIST}

For the MNIST handwritten digits recognition task, we implement a standard modern incarnation of the classic LeNet-5 architecture~\citep{lecun1998gradient}, with maxpooling, ReLU activations and dropout after the dense hidden layer. The total parameter count of the used baseline model is 61706. On the full MNIST dataset, this model achieves $99.1\%$ test accuracy without data augmentation.

We train our models using Adam optimizer ($\beta_1=0.9$, $\beta_2=0.999$) with a batch size of $50$. The learning rate is $0.1$, which is divided by a factor of $10$ at $50\%$ and $75\%$ of the training iterations. We use a weight decay of $0.0005$ and a dropout rate of $0.5$. Training is stopped after 10000 minibatches. 
%We have verified that this is enough to achieve convergence for all of our scenarios. Moreover, early stopping does not provide measurable improvements, as the test accuracy curve plateaus without later accuracy degradation.
$10000$ points from the training set are set aside as a held-out development set for hyperparameter tuning.

\subsection{CIFAR-10, CIFAR-100}

We use a residual network (ResNet) architecture~\citep{DBLP:journals/corr/HeZRS15} for the CIFAR-10 and CIFAR-100~\citep{Krizhevsky09} classification tasks, which consists of three levels, each containing stacked residual blocks. The structure of a residual block is conv-batchnorm-ReLU-conv-batchnorm, followed by elementwise addition and a ReLU nonlinearity. The three levels differ only in the number of filters and feature map sizes, which are 48, 96 and 192, respectively. The network has around $2.5$ million parameters. On the full (augmented) CIFAR-10 training set, this model achieves $93.71\%$ test accuracy.

The models are trained for 50000 iterations using SGD, with a momentum of $0.9$ and Nesterov momentum with dampening of~$0$. Mini-batch size is $128$. The initial learning rate is $0.1$, which is divided by a factor of $10$ at $50\%$ and $75\%$ of the training iterations. Weight decay is $0.003$.

\subsection{TinyImageNet-200}

We use the ResNet-18 architecture that has been used for the ImageNet training in \citet{DBLP:journals/corr/HeZRS15} with the appropriate modification for 200 labels: the output dimensions of the fully connected layer at the top of the network is 200. The filter numbers for the four levels are 64, 128, 256, 512. The network has around $11.3$ million parameters.
%On the full TinyImageNet-200 training set, this model achieves $\%$ test accuracy.

The models are trained for 75 epochs using Adam optimizer with parameters $\beta_1=0.9, \beta_2=0.999, \epsilon=10^{-8}$. Mini-batch size is $64$. The initial learning rate is $0.001$, which is divided by a factor of $10$ at the end of the $25th$ and $50th$ training epochs. Weight decay is set to $0.0001$ for the baseline model.

\subsection{SIN}
We use a synthetic dataset \emph{SIN} generated from the function $f(x) = \sin(5x)$ which allows for visualizing the function learned by the neural network. We use a small training set of $100$ points sampled uniformly from $[-1,1]$ and add some Gaussian noise ($\sigma=0.1$) to the output. For evaluation, we use a fixed grid of $900$ points from the same $[-1, 1]$ domain with uncorrupted outputs. The added noise highlights the need for regularization since overfitting on the training set can severely damage performance on the test set.

As a baseline model, we use an MLP network with 5 dense layers of $64$ neurons and ReLU nonlinearity, followed by a single linear neuron that produces the output of the network.

\section{Experiment specific hyperparameters}
Here we provide the optimal weights of the various regularizers in our experiments.

\subsection{4.1}
On MNIST, without weight decay, optimal weights are: SpectReg $=0.03$, Datagrad $=50$. With weight decay ($0.0005$), optimal weights are: SpectReg $=0.05$, DataGrad $=50$.

On CIFAR-10, without weight decay, optimal SpectReg weight is $0.001$. With weight decay ($0.0005$), optimal SpectReg weight is $0.03$.

On CIFAR-100, without weight decay, optimal SpectReg weight is $0.001$. With weight decay ($0.003$), optimal SpectReg weight is $0.0003$. %$0.003$.

\subsection{4.2}
Optimal DoubleBack/SpectReg weights are: Batchnorm $0.001/0.001$, Dropout $50/0.01$, Baseline $50/0.01$.

\subsection{4.3}
Optimal weights on MNIST for DoubleBack, SpectReg and CP are $50$, $0.03$ and $0.01$, respectively, and when CP and DoubleBack and used in conjunction, the optimal DoubleBack weight is $10$.

Optimal weights on CIFAR-10 for CP, SpectReg and DoubleBack are $0.003$, $0.03$ and $1$, respectively.

Optimal weights on CIFAR-100 for CP, SpectReg and DoubleBack are $0.00001$, $0.00003$ and $0.003$, respectively.

\subsection{4.4}
In Table~\ref{tab:mnist_trainsize} we give the weight of each regularizer for each training set size.

\begin{table}[htb]
    \caption{
        Optimal weights for each regularizer and each training set size.
    }
    \centering
    \begin{tabular}{l l l l l }
        Train size & DoubleBack & SpectReg & JacReg & CP \\
        \hline
        500 & 50 & 0.03 & 0.3 & 0.01 \\
        1000 & 50 & 0.03 & 0.03 & 0.01 \\
        2000 & 50 & 0.03 & 1 & 0.01 \\
        3000 & 20 & 0.03 & 1 & 0.01 \\
        4000 & 20 & 0.03 & 1 & 0.01 \\
        5000 & 20 & 0.003 & 1 & 0.1 \\
        10000 & 5 & 0.01 & 1 & 0.1 \\
        15000 & 2 & 0.01 & 1 & 0.01 \\
        20000 & 2 & 0.001 & 0.3 & 0.03
    \end{tabular}
    \label{tab:mnist_trainsize}
\end{table}

\subsection{4.5}

On TinyImageNet-200 we use SpectReg $=20000.0$. When used, weight decay is set to $0.0001$.

\subsection{4.6}
Optimal weights for SpectReg, JacReg and FrobReg are $0.03$, $1$ and $0.03$, respectively.

%\subsection{4.7} % smart WD experiment

%\subsection{4.8} % SIN experiment

\section{Gradient Regularization's effect on the magnitude of network parameters}

\begin{figure}[htb]
    \centering
    \includegraphics[width=\textwidth]{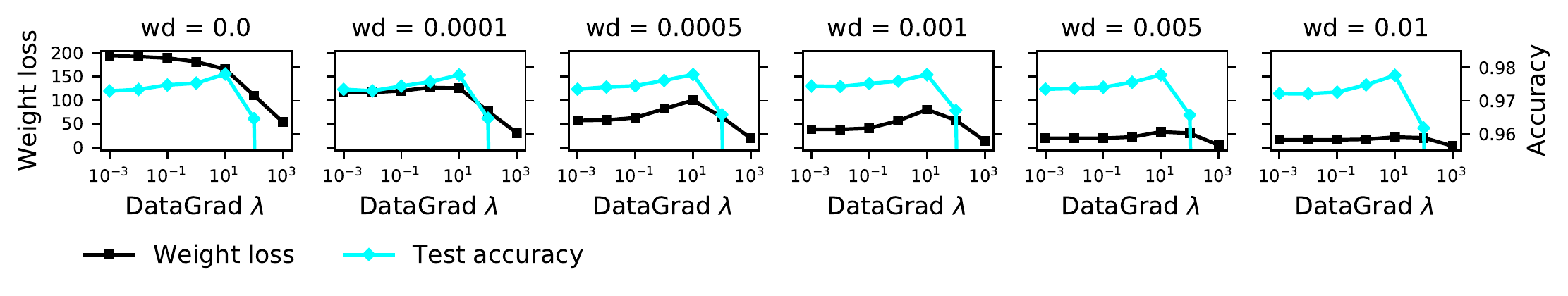}
    \caption{Black: weight loss, primary y-axis. Blue: accuracy, secondary y-axis. Each chart uses different weight decay. Increasing $\lambda$ increases accuracy, until a phase transition where accuracy collapses. Below this transition point, the weight loss of the trained network increases with $\lambda$, except for the case of zero weight decay, where DoubleBack takes the role of weight decay.
    }
    \label{fig:datagrad_vs_weight_loss}
\end{figure}

Figure~\ref{fig:datagrad_vs_weight_loss} shows the effect of DoubleBack weight $\lambda$ on accuracy, and on the squared L2 weight norm. There is a positive correlation between $\lambda$ and accuracy until a phase transition point where accuracy collapses. For small MNIST, this phase transition point is around DoubleBack $\lambda = 20$, regardless of the amount of weight decay.

Without weight decay, DoubleBack acts as a kind of weight decay itself, inasmuch as increasing the DoubleBack weight decreases the weight loss of the trained network. For models with weight decay, the relationship is more complex: below the phase transition, DoubleBack counteracts weight decay, above the phase transition it reinforces its effect.

The fact that DoubleBack affects a global attribute of the network such as total weight loss supports the hypothesis that its effect is global, not limited to the immediate neighborhood of the training points where the gradient norm is regularized. In the next subsection we give a clear demonstration of this phenomenon on our synthetic task.

\end{appendices}

\end{document}